\documentclass[12pt]{article}
\usepackage{fullpage}

\usepackage{amsmath,amssymb}
\usepackage{graphicx}
\usepackage{algorithm}
\usepackage{algpseudocode}
\usepackage{tabularx}
\usepackage{multirow}
\usepackage{xcolor}
\usepackage{fancyvrb}

\usepackage{pgfplots}
\pgfplotsset{compat=1.18}

\usepackage{tikz}
\usetikzlibrary{
    positioning,
    arrows.meta,
    shapes.geometric,
    fit
}

\DefineVerbatimEnvironment{Highlighting}{Verbatim}{commandchars=\\\{\}}

\begin{document}

\title{Learning to Access Computation: Accessibility Plasticity as a
Principle of Adaptive Intelligence}

\author{
  Zhaowen Fan
  \thanks{
    Department of Computer Science, Sichuan University,
    Email: \texttt{iamsuperfan@stu.scu.edu.cn}
  }
}
\date{}

\maketitle

\begin{abstract}
Modern neural networks primarily adapt through parameter modification within
predefined computational structures. While recent methods introduce modularity,
conditional computation, and parameter-efficient adaptation, they generally do
not distinguish computational capability from computational accessibility as
separate adaptive variables. This work introduces Accessibility Plasticity, a
principle of adaptive computation in which systems adapt not only by changing
what computation exists, but also by reorganizing which existing computations
can interact and participate. We formalize Accessibility Plasticity through a
relationship-based operational realization and establish a reuse-first hierarchy of
adaptation, where accessibility modification precedes more costly capability and
structural changes. A proof-of-concept evaluation on sequential learning tasks
shows that accessibility adaptation can reduce capability modification while
maintaining comparable task performance. These results suggest accessibility as
a distinct adaptive dimension and provide a foundation for future dynamic neural
systems whose computational relationships evolve with changing environments.
\end{abstract}

\section*{Keywords}

Accessibility Plasticity; adaptive computation; continual learning; neural
plasticity; conditional computation; dynamic neural systems

\section{Introduction}

Most contemporary artificial neural systems are optimized under a
parameter-centric learning paradigm: a computational architecture is specified
before deployment, and adaptation is primarily achieved by modifying numerical
parameters within that fixed computational organization \cite{lecun2015deep}. This paradigm has
enabled remarkable progress in large-scale perception, language modeling, and
decision-making, with modern foundation models achieving unprecedented
capabilities through large-scale pretraining and parameter optimization
\cite{vaswani2017attention, brown2020language, mnih2015human,
bommasani2021opportunities}. However, when deployed in dynamic environments
with continuously evolving data distributions, tasks, and operational conditions,
parameter-centric adaptation remains costly and prone to catastrophic forgetting,
limiting the ability to preserve and efficiently reuse previously acquired
capabilities \cite{parisi2019continual}.

Recent advances have explored alternatives by introducing modularity,
\cite{pfeiffer2023modular} conditional computation \cite{bengio2013estimating},
and parameter-efficient adaptation \cite{ding2023parameter}. Residual
architectures such as ResNet \cite{he2016deep} demonstrate the value of
preserving existing computational pathways, while graph neural networks
\cite{kipf2016semi, battaglia2018relational} highlight the role of
relationships among computational units in information propagation. Adapters
\cite{houlsby2019parameter} and low-rank adaptation \cite{hu2022lora} reduce
modification cost by restricting parameter updates, whereas mixture-of-experts
\cite{shazeer2017outrageously}, attention mechanisms \cite{vaswani2017attention},
and routing strategies \cite{rosenbaum2017routing} dynamically regulate
information flow and component selection. Nevertheless, these approaches are
generally developed as architecture-specific mechanisms rather than as a unified
adaptive principle. They manipulate connectivity, parameter subsets, expert
selection, or information routing, but do not explicitly distinguish
computational capability from computational accessibility or treat the evolving
relationships among computations as a first-class adaptive object.

Biological intelligence suggests that such a separation may be fundamental.
\cite{hassabis2017neuroscience} Organisms rarely respond to environmental changes
by immediately reconstructing their entire computational machinery; instead,
adaptation occurs across multiple timescales \cite{richards2019deep}. Existing
pathways are reused whenever possible \cite{anderson2010neural}, functional
relationships and activity patterns are reorganized, synaptic strengths are
modified \cite{magee2020synaptic}, and only over longer timescales does
large-scale structural reorganization occur \cite{holtmaat2009experience}. This
ordering suggests that efficient adaptation may depend not only on learning new
computations, but also on dynamically controlling access to existing
computational resources \cite{marder2002cellular}, motivating the concept of
Accessibility Plasticity introduced in this work.

This paper investigates the following question:

\textit{Can artificial systems adapt more efficiently by modifying how existing
computation is accessed before modifying what computation exists?}

We introduce Accessibility Plasticity as a distinct adaptive degree of freedom
that separates computational capability from computational accessibility: the
former describes what transformations a system can perform, while the latter
describes which computational components can interact during adaptation. Under
this perspective, learning is not only a process of changing computational
operators, but also a process of reorganizing relationships among existing
computations.

We formalize this principle through a relationship-based computational system in which
capability and accessibility evolve as separate states. The proposed reuse-first
principle prioritizes accessibility adaptation before more costly capability or
structural modifications. We show through a proof-of-concept evaluation that
accessibility adaptation can reduce capability modification while maintaining
comparable task performance in sequential learning scenarios. The operational
mechanism studied in this work is one realization of Accessibility Plasticity,
providing a foundation for future adaptive systems in which computational
relationships evolve continuously with changing environments.

\section{Conceptual Framework}

\subsection{Capability, Accessibility, and Scope}

Standard feedforward descriptions of neural computation can be summarized as

$$
\text{Architecture} \rightarrow \text{Parameters} \rightarrow \text{Computation}.
$$

In that view, the architecture determines which transformations may interact, while
the learned parameters determine the specific behavior realized along those fixed
routes.

The proposed perspective introduces an additional dimension. Architecture does not
fully determine realized computation; it provides only potential computation.
Accessible computation depends on the relationships that determine whether existing
capabilities can interact under current conditions.

The adaptive computational substrate is the formal object used here to represent
this idea. It is not the claim itself. The claim is that accessibility should be
treated as a distinct computational object, alongside capability, and that its
adaptation can precede modification of computational capability.

At the most compact level, the manuscript distinguishes two fundamental
computational objects:

$$
\begin{aligned}
W &\quad \text{determines what computation exists,} \\
\mathcal{A} &\quad \text{determines what computation is accessible.}
\end{aligned}
$$

The theory concerns the second object without reducing it to the first.

This manuscript does not claim to provide a complete theory of intelligence. It
proposes a missing adaptive degree of freedom: the ability of a computational
system to modify accessibility independently from capability. The purpose of the
paper is therefore to isolate and formalize that degree of freedom, not to claim
that all adaptation reduces to this single mechanism.

\subsection{Reuse Before Modification}

The conceptual principle is:

\begin{quote}
\emph{
An intelligent adaptive system should exhaust cheaper and more reversible forms of
modification before invoking more expensive and more persistent ones.
}
\end{quote}

This principle changes the optimization target. Traditional optimization primarily
focuses on maximizing task performance. The proposed paradigm instead seeks
performance-efficient adaptation:

\begin{equation}
\label{principle}
\max \;
\frac{
P(\mathcal{T}) \cdot R(\mathcal{S})
}{
C_{\mathrm{adapt}}
}
\end{equation}

where $P(\mathcal{T})$ denotes task performance, $R(\mathcal{S})$ denotes
computational reuse of the substrate, and $C_{\mathrm{adapt}}$ denotes the cost
of changing the system.

The shift is conceptual as well as technical: intelligence is not only the ability
to achieve a goal, but the ability to achieve goals by efficiently reusing existing
computational resources. This formulation is particularly relevant to continual
learning, where repeated parameter overwriting can become a liability.

\subsection{Accessibility Plasticity versus Selection}

Selection answers which already accessible path should be used now.
Accessibility plasticity answers how the future structure of accessible
computation should change through experience. The first is an inference-time
choice inside a given accessibility state; the second is a learning-time
modification of that accessibility state itself.

This distinction matters because a selection mechanism can choose among already
available possibilities without changing the substrate that makes those
possibilities available. Accessibility plasticity instead changes the accessibility
landscape so that future computations are shaped by past experience.
Thus accessibility is not only a parameter to be optimized at a single endpoint.
It can be treated as a dynamically evolving computational state whose trajectory
reflects interaction with a changing environment.

\section{Formalization}

\subsection{Adaptive Computational Substrate}

We represent the substrate as an undirected graph

\begin{equation}
G = (V, E).
\end{equation}

Each node $v_i \in V$ is a computational unit

\begin{equation}
v_i = W_i,
\end{equation}

where $W_i$ denotes the local transition operator and no persistent node state is
stored inside the vertex itself. For a transient input signal $x_i^t$, the node
computes

\begin{equation}
\tilde{x}_i^{\,t} = W_i x_i^t,
\end{equation}

so the node contributes only its transition matrix and does not introduce an
additional nonlinear transformation at this stage.

Each undirected relation index $e_{ij}=e_{ji}\in E$ marks a potential
computational relationship. The adaptive computational object associated with that
potential relationship is the relationship state:

\begin{equation}
R_{ij}=(\mathcal{F}_{ij},d_{ij}),
\end{equation}

where $\mathcal{F}_{ij}$ is a learned compatibility function and $d_{ij}\geq0$
denotes computational resistance. Given an information flow, the relationship state
evaluates the compatibility of the interaction and modifies the future
accessibility of this computational relationship. The evolving relationship state
already encodes the history of previous interactions, so no separate relation-memory
variable is required.

\subsection{Relationship State and Accessibility}

Computation in an adaptive computational substrate is not determined solely by the
existence of computational units or by fixed connections between them. Instead,
relationships between computational units define an adaptive accessibility
structure that determines which transformations can participate in the current
computation.

Crucially, accessibility is not identical to the relation set $E$. The set $E$
specifies potential relationships in the substrate, whereas accessibility
specifies which relationships become computationally available under the current
adaptive state. Relationship states are therefore one realization of
accessibility, not accessibility itself.

For a potential relationship between computational units $i$ and $j$, the
relationship state evaluates an interaction tendency:

\begin{equation}
s_{ij}^{t}
=
\mathcal{F}_{ij}(x_i^t,x_j^t),
\end{equation}

where $\mathcal{F}_{ij}$ represents a discrimination mechanism that evaluates
whether interaction between the two computational capabilities is
computationally meaningful.

The tendency is modulated by a resistance term:

\begin{equation}
z_{ij}^{t}
=
s_{ij}^{t}-\beta d_{ij},
\end{equation}

where $d_{ij}\geq0$ represents the intrinsic difficulty of establishing the
relationship and $\beta$ controls the influence of resistance.

The accessibility of the relationship is then determined by

\begin{equation}
a_{ij}^{t}=g(z_{ij}^{t}),
\end{equation}

where $g(\cdot)$ maps the evaluated tendency into an accessibility state. The
resulting accessibility does not represent a message weight. Instead, it defines
whether and to what degree the computational relationship becomes available
within the current substrate configuration.

The effective computational substrate at time $t$ is therefore induced by the
set of accessible relationships:

\begin{equation}
E_t^{*}
=
\{(i,j)\in E \mid a_{ij}^{t} > \theta\},
\end{equation}

which forms a dynamically accessible substrate:

\begin{equation}
G_t=(V,E_t^{*}).
\end{equation}

Multiple adaptive relationships may become simultaneously accessible, producing
concurrent computational pathways within the induced substrate $G_t$.
Accordingly, the system does not rely on a single predetermined execution
trajectory. Instead, computation emerges from the collective configuration of
active relationships.

The central distinction is that relationships are not passive communication
channels. Their states are adaptive computational structures that evaluate, enable,
suppress, and reshape future accessibility between existing capabilities.

At the principle level, accessibility evolves according to an abstract update
operator:

\begin{equation}
\mathcal A_{t+1}
=
\mathcal P_A(\mathcal A_t,\mathcal X_t).
\end{equation}

This equation is intentionally more general than the particular relationship-state
implementation used in the proof-of-concept experiments. It states that
accessibility is a time-indexed computational object whose future state may depend
on its present configuration and on the encountered environment $\mathcal X_t$.
Updates to $\mathcal F_{ij}$ or $d_{ij}$ are possible realizations of this
principle, not definitions of Accessibility Plasticity itself.

\subsection{Computational Interpretation}

The model separates two usually entangled concepts:

\begin{itemize}
\item node operators determine \emph{what} local computation can be performed;
\item accessibility states determine \emph{whether and how} existing computation
can be reached.
\end{itemize}

This separation is the core design move of the paradigm. Learning is therefore not
only the acquisition or modification of computational capability. It is also the
reorganization of access to capabilities that already exist in the substrate. The
system can respond to new tasks by reshaping relationships to existing operators
before rewriting the operators themselves.

Unlike conventional neural networks, where learning is primarily stored in node
parameters, the proposed substrate allows learning to emerge from the evolution of
relationships between computational units. Adaptation occurs through the evolution
of computational relationships, not through explicit storage of interaction
history.

In compact form, the interpretation is:

$$
\begin{aligned}
W &\Rightarrow \text{what computation exists}, \\
\mathcal{A} &\Rightarrow \text{what computation is accessible}.
\end{aligned}
$$

The contribution of the paper is Accessibility Plasticity as a first-class object
of learning, not a new family of network architectures.

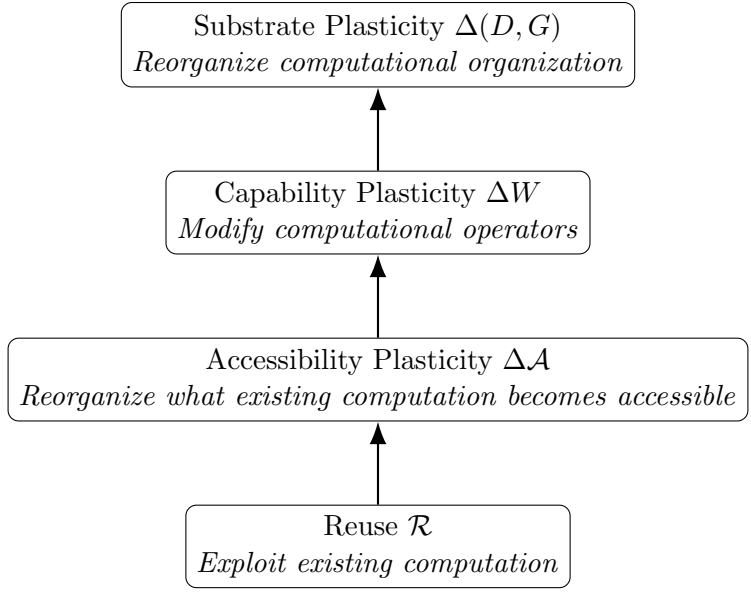
\begin{figure}[t]
\centering
\begin{tikzpicture}[
    node distance=1.1cm,
    levelbox/.style={
        rectangle,
        rounded corners,
        draw,
        minimum width=4.9cm,
        minimum height=0.95cm,
        align=center,
        font=\small
    },
    note/.style={
        align=left,
        font=\small
    },
    arrow/.style={
        -{Latex[length=3mm]},
        thick
    }
]

\node[levelbox] (reuse)
{Reuse $\mathcal{R}$\\
\textit{Exploit existing computation}};

\node[levelbox, above=of reuse] (access)
{Accessibility Plasticity $\Delta\mathcal{A}$\\
\textit{Reorganize what existing computation becomes accessible}};

\node[levelbox, above=of access] (capability)
{Capability Plasticity $\Delta W$\\
\textit{Modify computational operators}};

\node[levelbox, above=of capability] (substrate)
{Substrate Plasticity $\Delta(D,G)$\\
\textit{Reorganize computational organization}};

\draw[arrow] (reuse) -- (access);
\draw[arrow] (access) -- (capability);
\draw[arrow] (capability) -- (substrate);

\end{tikzpicture}
\caption{
Adaptive systems possess multiple degrees of adaptive freedom. The proposed
hierarchy orders forms of adaptation by increasing depth: systems should
preferentially reuse existing computation, then modify computational
accessibility, then modify computational capability, and only at the deepest
level reorganize computational organization itself. Lower levels preserve more
of the current computational identity of the system, whereas higher levels incur
greater modification cost and structural persistence. The hierarchy denotes an
ordering of adaptive depth rather than a mandatory optimization schedule.
}
\label{fig:adaptive_hierarchy}
\end{figure}

\subsection{Hierarchy of Adaptive Depth}

Adaptive systems possess multiple degrees of adaptive freedom. These degrees of
freedom differ in persistence, reversibility, and modification cost. Adaptive
freedom refers to the hierarchy of internal degrees of freedom through which a
system can respond to environmental change. We propose a reuse-first ordering in
which a system first exploits existing computational capability, then reorganizes
computational accessibility, then modifies computational capability itself, and
only finally reconfigures the organization of computation.

The hierarchy can be summarized as:

$$
\mathcal{R}
\prec
\Delta \mathcal{A}
\prec
\Delta W
\prec
\Delta(D,G)
$$

where $\mathcal{R}$ denotes pure computational reuse without modification of
accessibility, capability, or organization; $\Delta \mathcal{A}$ denotes
changes in computational accessibility; $\Delta W$ denotes modifications of
computational capability; and $\Delta(D,G)$ denotes reorganization of the
computational substrate.

The ordering does not represent a fixed algorithmic procedure or a mandatory
optimization sequence. It is an ordering of adaptation depth. Lower levels
preserve more of the existing computational identity of the system, whereas higher
levels introduce increasingly fundamental changes to what the system can compute
and how computation is organized.

\paragraph{Reuse}

The first adaptive response is to exploit existing computational capability
without modification. Given a new task or a shift in input distribution, the
system first evaluates whether its current computational organization already
contains sufficient capability to solve the problem.

Successful reuse preserves existing knowledge and incurs minimal adaptation cost
because neither accessibility relations nor computational operators are altered.

\paragraph{Accessibility plasticity}

When the required computational capability already exists but is not effectively
reachable under the current configuration, adaptation occurs through changes in
computational accessibility. Accessibility is therefore not an additional
computational capability; it is a property describing whether existing
capabilities can participate in computation.

We refer to this process as accessibility plasticity. In the adaptive computational
substrate studied here, relationship-state evolution provides one substrate-level
realization of this principle. Unlike mechanisms that merely select among existing
pathways at inference time, accessibility plasticity modifies the underlying
relationship landscape through which future computational pathways emerge.

Let the accessibility configuration of the computational substrate be

\begin{equation}
\mathcal{A}_t=\{a_{ij}^{t}\},
\end{equation}

where $a_{ij}^{t}$ represents the accessibility of the computational
relationship between units $i$ and $j$ under the current substrate state.

For the adaptive computational substrate, accessibility is induced by the
relationship state $R_{ij}^{t}=(\mathcal{F}_{ij}^{t},d_{ij}^{t})$:

\begin{equation}
\label{equation}
a_{ij}^{t}
=
g
\left(
\mathcal{F}_{ij}^{t}(x_i^t,x_j^t)
-
\beta d_{ij}^{t}
\right),
\end{equation}

where $\mathcal{F}_{ij}^{t}$ evaluates relational compatibility and $d_{ij}^{t}$
represents computational distance between units. The mapping from relationship
state to accessibility is not unique; the purpose of Eq.~\ref{equation} is to
provide one realizable coordinate system for an otherwise abstract adaptive variable.

The evolution of accessibility is therefore described at an abstract level as:

\begin{equation}
\mathcal{A}_{t+1}
=
\mathcal{P}_{A}
(\mathcal{A}_{t},\mathcal{X}_{t}),
\end{equation}

where $\mathcal{P}_{A}$ denotes an accessibility plasticity operator and
$\mathcal{X}_{t}$ represents the computational experience accumulated by the
system at time $t$, including interaction outcomes, task feedback, and
environmental signals. This formulation emphasizes that accessibility is itself
an adaptive state variable whose evolution is governed by experience, rather
than a fixed property of the computational substrate. The abstract update
$\mathcal{A}_{t+1}=\mathcal{P}_{A}(\mathcal{A}_{t},\mathcal{X}_{t})$ is therefore
the protected theoretical statement; any concrete distance-based update or
relationship-state update is only one possible realization, not the definition of
Accessibility Plasticity.

A concrete realization of this principle can be obtained by adapting the
underlying relationship geometry. For example, computational distance may evolve
according to:

\begin{equation}
d_{ij}^{t+1}
=
d_{ij}^{t}
-
\eta_d s_{ij}^{t},
\end{equation}

where $s_{ij}^{t}$ denotes the interaction outcome between computational units
$i$ and $j$. Successful interactions reduce computational distance and
increase future accessibility, whereas unsuccessful interactions weaken the
corresponding relationship. In this realization, the history of interaction is
implicitly stored in the evolving relationship state rather than through an
external memory mechanism.

Therefore, adaptation occurs first by reshaping access to existing computation
rather than modifying the computation itself.

\paragraph{Capability plasticity}

When accessibility adaptation is insufficient because the existing computational
capability cannot adequately represent the required transformation within the
current substrate, the system modifies its local computational capabilities.

This corresponds to conventional parameter adaptation, but with a stronger
interpretation: capability plasticity changes what computation is available within
the substrate, whereas accessibility plasticity changes whether existing
computation can be reached.

\begin{equation}
W_i
\leftarrow
W_i+\Delta W_i .
\end{equation}

Because modifying operators can overwrite previously acquired competence, it is
considered a deeper and more costly form of adaptation than accessibility
plasticity.

\paragraph{Substrate plasticity}

The highest level of adaptation occurs when both accessibility and capability
modification become insufficient. At this level, the organization of computation
itself changes.

The learned computational distance matrix $D=[d_{ij}]$ defines the relational
geometry of the substrate. Persistent low-distance regions indicate strong
computational affinity, while high-distance regions indicate weak relationships.

Through this learned geometry, the substrate may reorganize itself by:

\begin{itemize}
\item forming computational modules with persistent relational affinity;
\item dissolving obsolete structures when their internal relationships become
incompatible;
\item redistributing computational boundaries according to learned relational
geometry.
\end{itemize}

Therefore, substrate plasticity is not presented here as a search procedure over
architectures. It is the deepest adaptive level at which computational
organization itself changes.

A generic substrate adaptation objective can be written as:

\begin{equation}
\min_{D,G}
L_{\mathrm{task}}
+
\lambda_s C_s(D,G)
\end{equation}

where $C_s(D,G)$ measures the cost of reorganizing computational organization
under the learned relational geometry.

At this highest level, adaptation no longer changes only information flow or
individual computational operators. Instead, it transforms the computational
landscape in which future computation becomes possible.

\section{Operational Realization (Experimental Setup)}

The experiments are designed as proof-of-concept validation of the principle,
not as a benchmark campaign for state-of-the-art continual learning. Their role is
to test whether accessibility behaves as a distinct adaptive variable and whether
reuse-first adaptation can reduce shared capability modification under sequential
tasks. Concretely, the experimental design evaluates the following hypotheses:
\begin{itemize}
\item accessibility plasticity constitutes a meaningful adaptive degree of freedom;
\item adaptive systems can reduce capability modification by first modifying accessibility;
\item accessibility adaptation is not equivalent to merely adding extra adaptive parameters;
\item accessibility alone is insufficient for continual adaptation, consistent with the proposed hierarchy
$\mathcal{R} \prec \Delta \mathcal{A} \prec \Delta W \prec \Delta(D,G)$.
\end{itemize}

\subsection{Proof-of-Concept Realization and Training Protocol}

The reuse-first hierarchy defines a general preference over modification depth:
adaptation should first exploit existing computation, then reorganize
accessibility, and only afterwards rewrite computational capability. At the
principle level, the objective is not only low task loss, but low task loss with
minimal internal transformation. One practical realization therefore augments the
task objective with separate costs for accessibility and capability change:

\begin{equation}
\mathcal{L}
=
L_{\mathrm{task}}
+
\lambda_A C_A
+
\lambda_W C_W .
\end{equation}

The key point is not regularization by itself. The point is that the optimization
distinguishes modification depths by assigning different costs to changes in
accessibility and changes in capability. In the present instantiation,

\begin{equation}
\begin{aligned}
C_A
&=
\|\Delta \mathcal{F}\|_1
+
\|\Delta D\|_1,
\\
C_W
&=
\sum_i
\|\Delta W_i\|_F .
\end{aligned}
\end{equation}

The first term measures reorganization of relationship state, while the second
measures changes to shared computational operators. The intended inductive
preference is

$$
\lambda_A \leq \lambda_W,
$$

so that capability plasticity remains available but is invoked only when shallower
adaptation is insufficient.

Algorithm~\ref{alg:reuse_first} summarizes one operational realization of the
principle. It should be read as an instantiation rather than as the contribution
itself. The experiments implement the hierarchy using a performance-based
controller: at each task, the system attempts the shallowest adaptation depth and
avoids deeper modification if the current depth satisfies a fixed training
criterion. Operationally,

\begin{equation}
s^\ast
=
\min_s
\{s:\mathrm{Acc}(s)\geq \tau\},
\qquad
s\in\{\mathrm{reuse},\Delta\mathcal{A},\Delta W\}.
\end{equation}

Here $\tau=0.9$ is the task-sequential experimental performance constraint, not
the theoretical definition of Accessibility Plasticity. The dynamic datastream
demonstration uses a slightly stricter target-environment controller threshold
($\tau=0.92$) because it tracks progress toward the hardest stream endpoint
rather than per-task adaptation alone. The task-sequential experiments contain
three controlled stages for each task after the first:
\begin{enumerate}
\item \textbf{Reuse:} adapt only the task-specific readout head while freezing
shared capability and accessibility;
\item \textbf{Accessibility plasticity:} freeze shared capability and update only
the relationship state that induces accessibility;
\item \textbf{Capability plasticity:} freeze the learned relationship state and
update only the shared computational operators if the previous stages do not
achieve the task-accuracy threshold.
\end{enumerate}

The first task is used to initialize the shared capability bank, so it necessarily
invokes capability plasticity. Thereafter, adaptation-stage decisions are made on
the current task's training split, while the reported continual-learning metrics
are measured on held-out test data. This design avoids using test performance to
decide when deeper modification stages are activated. In the dynamic datastream
demonstration, the same staged preference is preserved, but reuse corresponds to
evaluating the current model state on the incoming stream slice rather than to
head-only adaptation.

\begin{figure}[t]
\centering
\begin{tikzpicture}[
  node distance=0.45cm and 0.9cm,
  block/.style={
    rectangle,
    rounded corners=3pt,
    draw,
    minimum width=3.2cm,
    minimum height=0.65cm,
    align=center,
    font=\scriptsize,
    inner sep=3pt
  },
  smallblock/.style={
    rectangle,
    rounded corners=3pt,
    draw,
    minimum width=2.7cm,
    minimum height=0.65cm,
    align=center,
    font=\scriptsize,
    inner sep=3pt
  },
  futureblock/.style={
    rectangle,
    rounded corners=3pt,
    draw,
    dashed,
    minimum width=2.7cm,
    minimum height=0.65cm,
    align=center,
    font=\scriptsize,
    inner sep=3pt
  },
  arrow/.style={
    -{Latex[length=2mm, width=1.3mm]},
    thick
  },
  dashedarrow/.style={
    -{Latex[length=2mm, width=1.3mm]},
    thick,
    dashed
  }
]


\node[block] (env)
{Environment and task input\\ $x_t,\mathcal X_t$};

\node[block, below=of env] (cap)
{Existing computational capability\\ $W_1,W_2,\dots,W_n$};

\node[block, below=of cap] (rel)
{Relationship state\\ $R_{ij}=(\mathcal F_{ij},d_{ij})$};

\node[block, below=of rel] (acc)
{Accessibility state\\ $\mathcal A_t$};

\node[block, below=of acc] (comp)
{Task computation and output\\ $y_t$};

\draw[arrow] (env) -- (cap);
\draw[arrow] (cap) -- (rel);
\draw[arrow] (rel) -- (acc);
\draw[arrow] (acc) -- (comp);


\node[smallblock, below=0.5cm of comp] (decision)
{Satisfied with the output?};

\draw[arrow] (comp) -- (decision);


\node[futureblock, right=of cap] (updateDG)
{Organizational plasticity\\ Update organization\\ $\Delta(D,G)$};

\node[smallblock, right=2cm of rel] (updateW)
{Capability plasticity\\ Update computation\\ $\Delta W$};

\node[smallblock, right=of acc] (updateA)
{Accessibility plasticity\\ Update relationship state $R_{t+1}$\\ $\mathcal A_{t+1}=\mathcal P_A(\mathcal A_t,\mathcal X_t)$};


\coordinate (bus1) at ([xshift=0.30cm, yshift=-5pt] updateA.east |- decision.east);
\coordinate (bus2) at ([xshift=0.55cm]              updateA.east |- decision.east);
\coordinate (bus3) at ([xshift=0.80cm, yshift=5pt]  updateA.east |- decision.east);

\draw[arrow] ([yshift=-5pt]decision.east) -- (bus1);
\draw[arrow] (decision.east)              -- (bus2);
\draw[dashedarrow] ([yshift=5pt]decision.east)  -- (bus3);

\draw[arrow] (bus1) |- (updateA.east);
\draw[arrow] (bus2) |- (updateW.east);
\draw[dashedarrow] (bus3) |- (updateDG.east);


\draw[arrow] (updateA.west) -- (acc.east);
\draw[arrow] (updateW.west) -- (cap.east);
\draw[dashedarrow] (updateDG.west) -- (cap.east);

\end{tikzpicture}

\caption{
Operational realization of the adaptive transformation hierarchy. The system
first evaluates output satisfaction. If adaptation is required, it proceeds
through a hierarchical routing structure: first modifying accessibility state
via relationship updates ($\mathcal A_{t+1}=\mathcal P_A(\mathcal A_t,\mathcal X_t)$),
then invoking capability plasticity ($\Delta W$) if needed, and finally engaging
theoretical organizational plasticity ($\Delta(D,G)$) as a deeper evolutionary layer.
}
\label{fig:operational_realization_overview}
\end{figure}
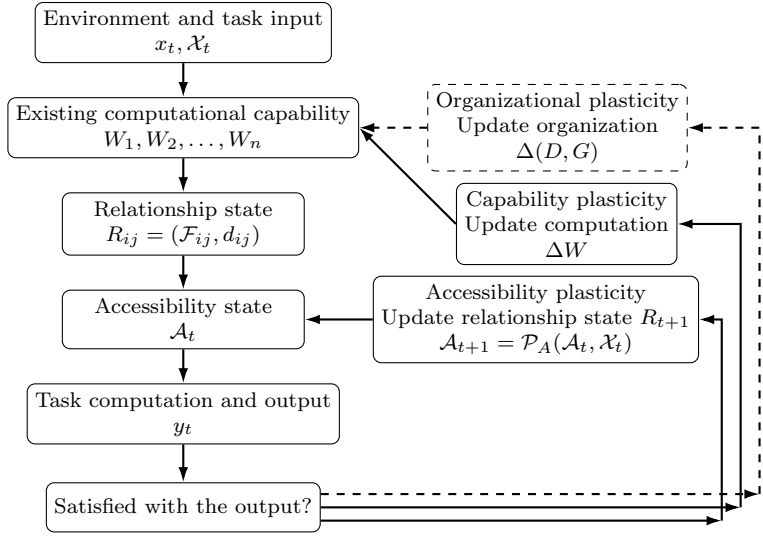

\begin{algorithm}[t]
\caption{One Reuse-First Instantiation in the Operational Realization}
\label{alg:reuse_first}
\begin{algorithmic}[1]

\Require computational graph $G=(V,E)$, task batch $\mathcal{B}$, adaptation thresholds
$\epsilon_r,\epsilon_A,\epsilon_W$

\State Evaluate current configuration on $\mathcal{B}$

\If{task loss $\leq \epsilon_r$}
    \State return without modification
\EndIf

\State Update accessibility state $\mathcal{A}$
\State Realize accessibility adaptation through relationship updates

\If{task loss $\leq \epsilon_A$}
    \State return after accessibility adaptation
\EndIf

\State Freeze accessibility state $\mathcal{A}$
\State Update computational operators $W$

\If{task loss $\leq \epsilon_W$}
    \State return after capability adaptation
\EndIf

\State Trigger deeper organizational reconfiguration if required

\end{algorithmic}
\end{algorithm}

\subsection{Datasets, Baselines, and Evaluation Metrics}

All experiments are CPU-only and use a single fixed random seed. Two controlled
continual-learning settings are evaluated.

\paragraph{Split MNIST.}
Five sequential binary tasks are constructed from MNIST:
$0$ versus $1$, $2$ versus $3$, $4$ versus $5$, $6$ versus $7$, and $8$ versus $9$.

\paragraph{Permuted MNIST.}
The same binary task sequence is retained, but each task receives a fixed
task-specific pixel permutation. This benchmark preserves the simple
continual-learning protocol while introducing a stronger distribution shift.

For both datasets, each class contributes $250$ training examples and $120$ test
examples per task. The baseline and EWC models are two-layer multilayer
perceptrons with hidden dimension $64$ and task-specific binary heads. The
operational realization of Accessibility Plasticity uses a shared bank of $8$
experts, hidden dimension $64$, and a task-conditioned relationship state
$R=(\mathcal{F},d)$ that induces accessibility over the experts. The experimental
comparisons include:
\begin{itemize}
\item \textbf{Baseline MLP:} direct capability modification of a shared trunk;
\item \textbf{Accessibility Plasticity Realization:} reuse-first accessibility plasticity with capability updates when necessary;
\item \textbf{Fixed Relationships:} the same capability bank with accessibility adaptation disabled;
\item \textbf{Accessibility Only:} accessibility adaptation allowed, later capability plasticity disabled;
\item \textbf{Random Accessibility Control:} the same accessibility-state budget as the operational realization, but accessibility updates are random rather than relationship-driven;
\item \textbf{EWC:} a classical weight-preservation baseline.
\end{itemize}

\subsubsection{Dynamic Accessibility Evolution}
In addition to the two task-sequential demonstrations, we include a lightweight
streaming distribution-shift experiment. Its purpose is not benchmark comparison,
but to test whether accessibility behaves as a dynamically evolving adaptive
state. A single binary MNIST classifier is first trained on an easy stationary
slice of digits $0$ versus $1$ to establish an initial computational state.
The same classifier and readout are then exposed to an eight-step datastream in
which permutation strength and noise level increase gradually over time. At each
stream step, the system first evaluates the current model state, then updates the
accessibility-inducing relationship state, and only then updates shared
capability parameters if accessibility adaptation remains insufficient. The main
readout is a learning-progress curve on the fixed hardest target distribution at
the endpoint of the stream, together with the measured accessibility movement
$\Delta A_t$ and shared capability movement $\Delta W_t$ at each step.

The primary reported metrics are final average accuracy, average forgetting,
capability modification cost $C_W$, accessibility modification cost $C_A$, and
modification-efficiency score

\begin{equation}
\mathrm{MES}
=
\frac{\mathrm{Performance}}
{C_A + C_W}.
\end{equation}

In the experiments, $C_W$ is measured as the accumulated movement of the shared
capability parameters only. For the baseline and EWC models, these parameters are
the shared MLP trunk. For the operational realization of Accessibility Plasticity,
they are the shared expert bank and shared post layer. Task-specific heads are
excluded from $C_W$, and relationship-state
parameters are excluded from $C_W$ and counted instead in $C_A$. This distinction
is important because the experimental claim concerns modification of shared
computational capability, not all trainable parameters indiscriminately.

\section{Results \& Discussion}

Table~\ref{tab:split_results} reports the results on Split MNIST, and
Table~\ref{tab:permuted_results} reports the results on Permuted MNIST. The
results should be interpreted as proof-of-concept evidence for the proposed
principle rather than as benchmark rankings.

\begin{table}[t]
\centering
\small
\begin{tabular}{lccccc}
\hline
Model & Accuracy & Forgetting & $C_W$ & $C_A$ & MES \\
\hline
Baseline MLP & 0.9583 & 0.0250 & 8.4460 & 0.0000 & 0.1135 \\
Accessibility Plasticity Realization & 0.9183 & 0.0177 & 2.8674 & 1.4131 & 0.2145 \\
Fixed Relationships & 0.9350 & 0.0052 & 6.8209 & 0.0000 & 0.1371 \\
Accessibility Only & 0.6625 & 0.0000 & 0.9107 & 1.2227 & 0.3105 \\
Random Accessibility Control & 0.8858 & 0.0083 & 1.8810 & 2.7926 & 0.1895 \\
EWC & 0.9658 & 0.0177 & 8.7388 & 0.0000 & 0.1105 \\
\hline
\end{tabular}
\caption{Proof-of-concept results on Split MNIST. The central comparison is not raw accuracy alone, but the relationship between performance and shared capability modification cost.}
\label{tab:split_results}
\end{table}

\begin{table}[t]
\centering
\small
\begin{tabular}{lccccc}
\hline
Model & Accuracy & Forgetting & $C_W$ & $C_A$ & MES \\
\hline
Baseline MLP & 0.9700 & 0.0125 & 9.5016 & 0.0000 & 0.1021 \\
Accessibility Plasticity Realization & 0.9242 & 0.0094 & 2.2484 & 0.5597 & 0.3291 \\
Fixed Relationships & 0.9408 & 0.0156 & 6.2389 & 0.0000 & 0.1508 \\
Accessibility Only & 0.6375 & 0.0000 & 0.9107 & 1.3141 & 0.2865 \\
Random Accessibility Control & 0.8608 & 0.0052 & 1.7982 & 2.9599 & 0.1809 \\
EWC & 0.9700 & 0.0104 & 9.3522 & 0.0000 & 0.1037 \\
\hline
\end{tabular}
\caption{Proof-of-concept results on Permuted MNIST. The qualitative pattern observed on Split MNIST remains visible under stronger distribution shift.}
\label{tab:permuted_results}
\end{table}

\begin{figure}[t]
\centering
\includegraphics[width=\linewidth]{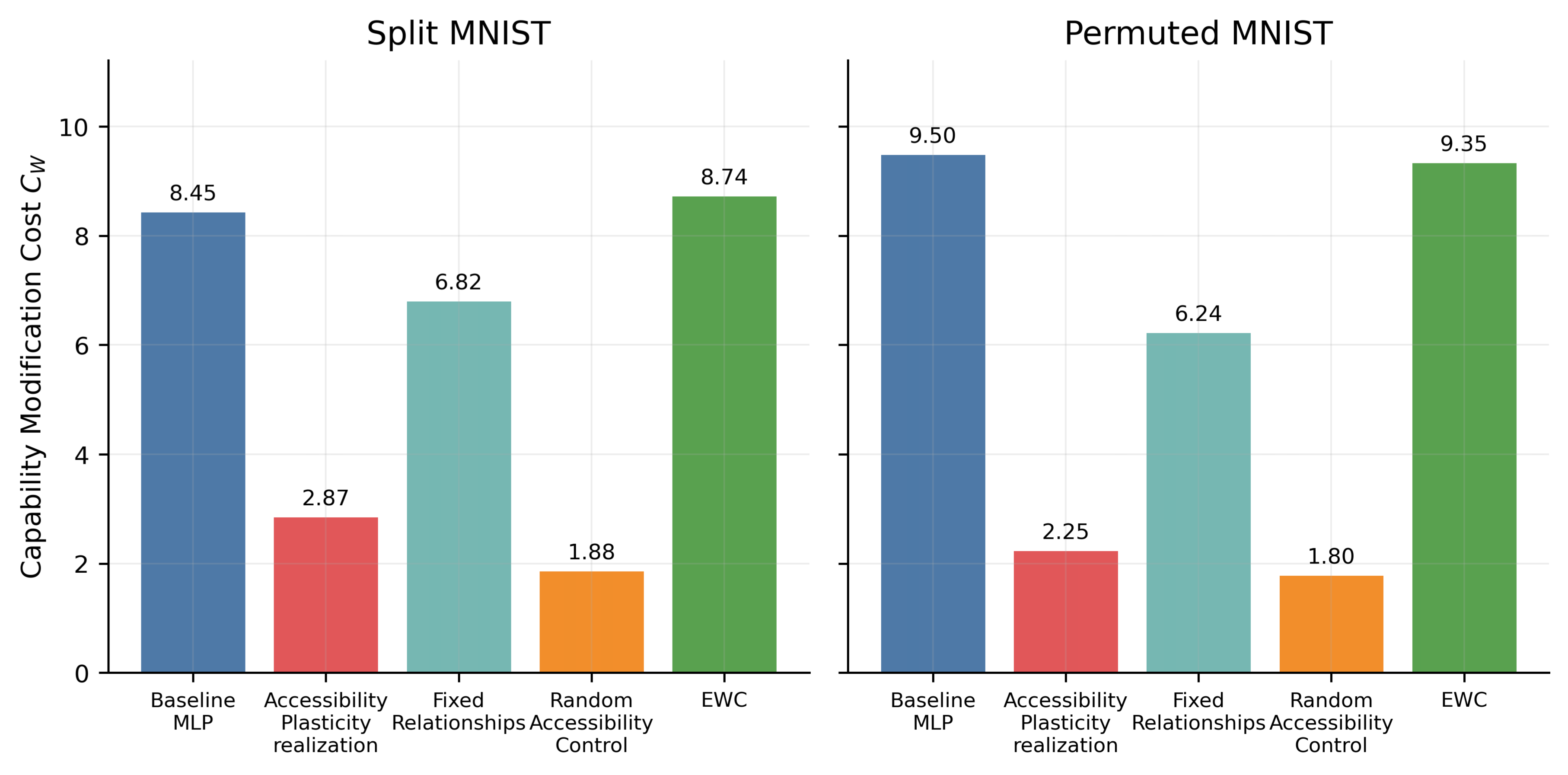}
\caption{
Capability modification cost across methods on Split MNIST and Permuted MNIST.
The operational realization of Accessibility Plasticity occupies a markedly lower
capability-modification regime than the direct weight-updating baselines while
remaining meaningfully above zero, indicating reduced but not eliminated
capability plasticity.
}
\label{fig:capability_cost}
\end{figure}

\begin{figure}[t]
\centering
\includegraphics[width=\linewidth]{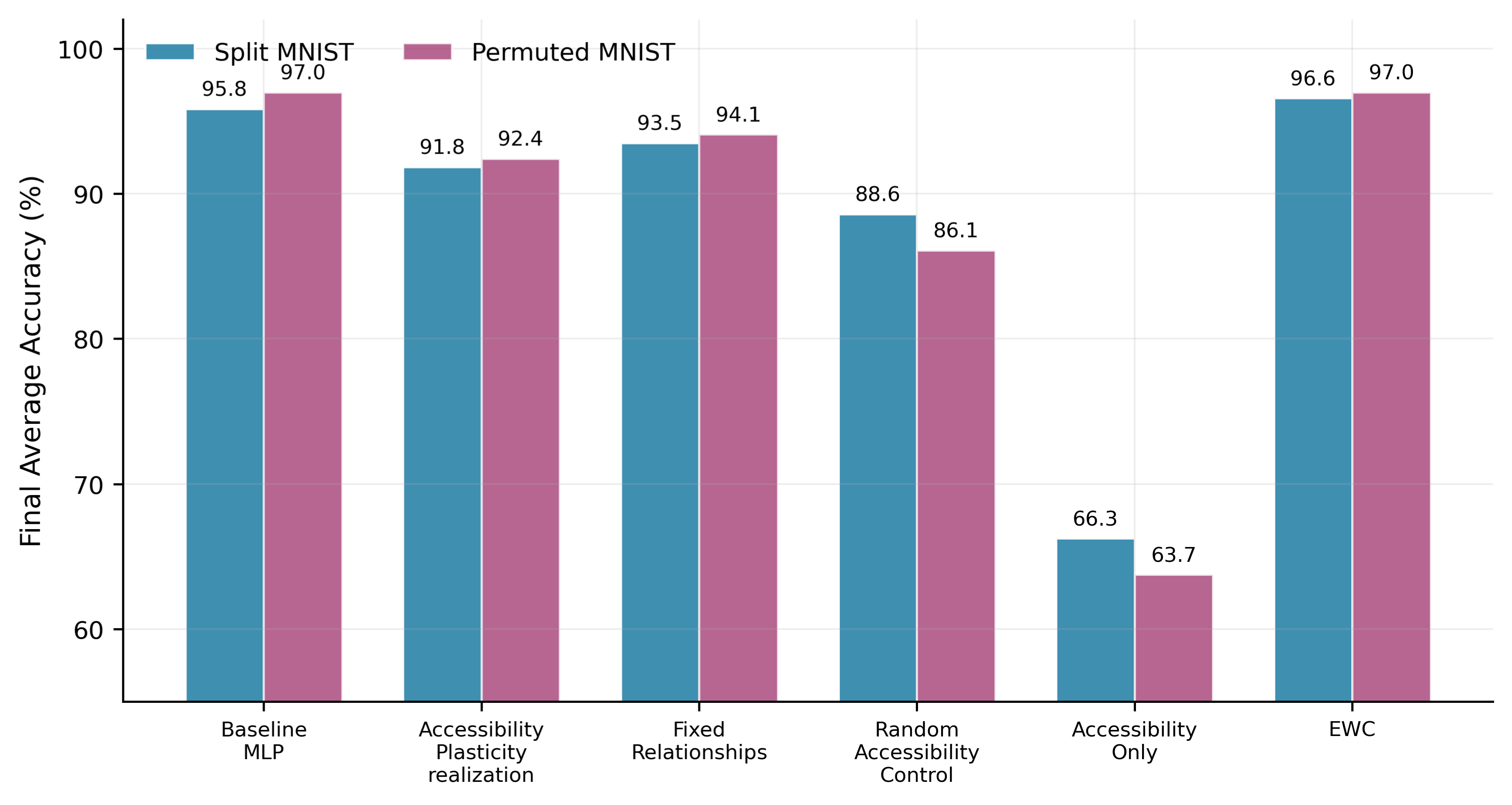}
\caption{
Final average accuracy on Split MNIST and Permuted MNIST. The purpose of the
comparison is not to identify the highest-accuracy method in isolation, but to
show that reduced capability modification under Accessibility Plasticity does not
come from collapse of task performance.
}
\label{fig:accuracy_comparison}
\end{figure}

\subsection{Experimental Observations}

The first empirical observation is that the operational realization of
Accessibility Plasticity does not achieve the highest raw accuracy. The strongest
raw accuracies are obtained by the baseline MLP and EWC. This is consistent with
the purpose of the experiments: they are not intended to establish state-of-the-art
continual learning, but to test whether adaptation can move into a lower
shared-capability-modification regime. Figures~\ref{fig:capability_cost} and
\ref{fig:accuracy_comparison} summarize this empirical trade-off directly.

\paragraph{Split MNIST.}
On Split MNIST, the operational realization of Accessibility Plasticity reaches
accuracy $0.9183$
with capability cost $2.8674$, compared with baseline accuracy $0.9583$ and
capability cost $8.4460$. The same model attains lower raw accuracy than Fixed
Relationships ($0.9350$), but requires much less capability modification
($2.8674$ versus $6.8209$). The Random Accessibility Control reaches
$0.8858$ accuracy while incurring accessibility cost $2.7926$, substantially higher
than the accessibility cost of the learned operational realization ($1.4131$).
The Accessibility Only ablation yields final accuracy $0.6625$, indicating that
accessibility adaptation alone does not carry the full sequential-learning burden.

\paragraph{Permuted MNIST.}
On Permuted MNIST, the same qualitative pattern remains visible under stronger
distribution shift. The operational realization reaches accuracy $0.9242$ with
capability cost $2.2484$, compared with baseline accuracy $0.9700$ and capability
cost $9.5016$. Fixed Relationships attains accuracy $0.9408$, but again with much
larger capability cost ($6.2389$). The Random Accessibility Control falls to
$0.8608$ accuracy while spending accessibility cost $2.9599$, compared with
$0.5597$ for the learned operational realization. Accessibility Only remains
insufficient, with final accuracy $0.6375$.

\begin{figure}[t]
\centering
\includegraphics[width=\linewidth]{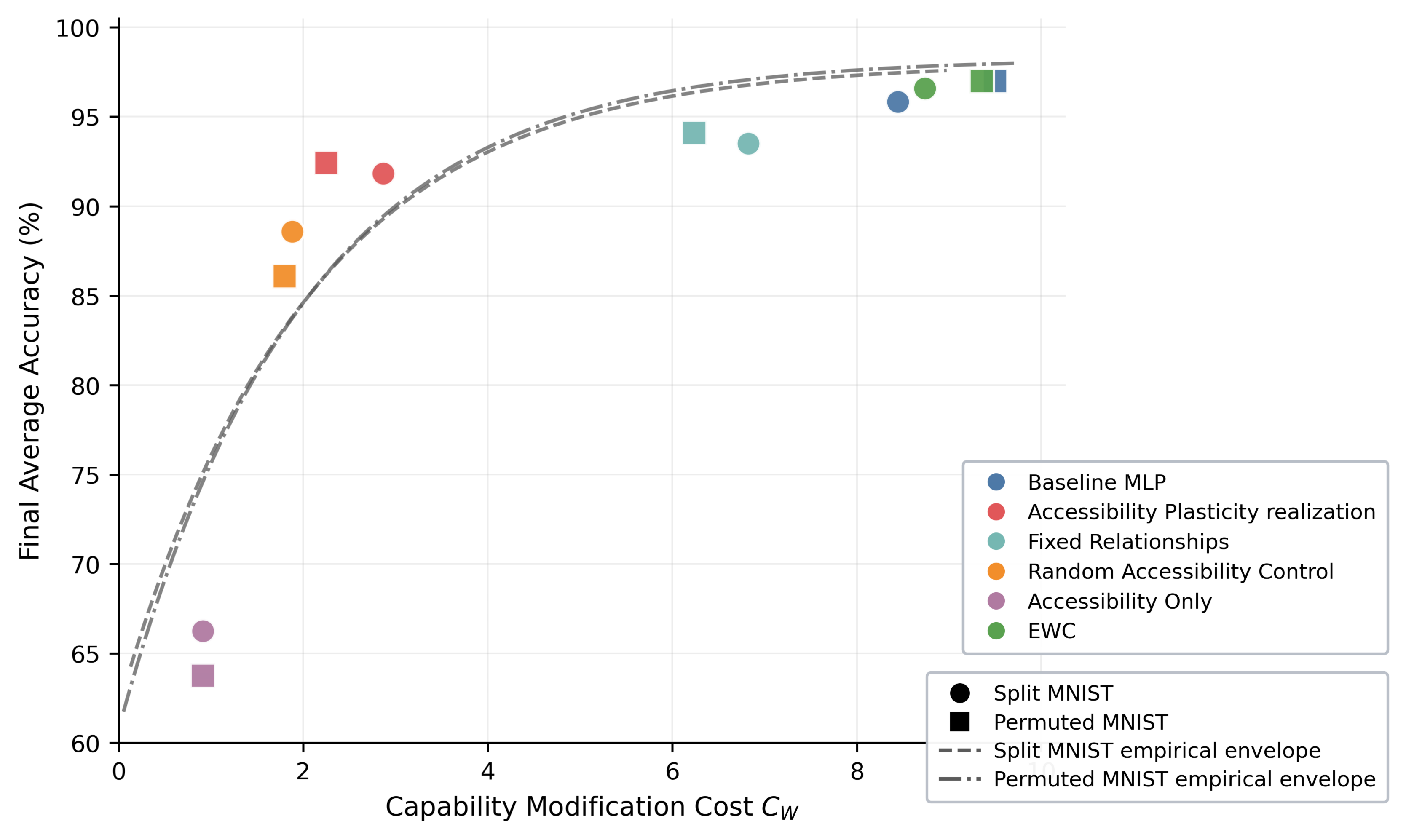}
\caption{
Accuracy-capability trade-off space for all methods on both datasets. Colors
identify methods and marker shapes identify datasets, allowing the figure to
compare adaptation regimes without crowding the plot with direct point labels.
Dashed and dash-dotted curves show smooth objective-front fits to the empirical
non-dominated trends within each dataset, not theoretical Pareto fronts over the
broader model class.
The operational realization of Accessibility Plasticity occupies a region with
competitive accuracy and substantially lower shared capability modification than
the direct weight-updating baselines.
}
\label{fig:tradeoff_space}
\end{figure}

\paragraph{Accuracy-capability trade-off.}
Figure~\ref{fig:tradeoff_space} summarizes the same empirical pattern in a single
view. The operational realization of Accessibility Plasticity remains well below
the baseline and EWC in capability modification cost while avoiding the severe
performance collapse of the Accessibility Only ablation. Fixed Relationships and
Random Accessibility Control occupy distinct intermediate regimes, reinforcing
that the central effect is not merely the presence of additional adaptive state,
but the learned use of accessibility as a separate adaptive degree of freedom.

\begin{figure}[t]
\centering
\includegraphics[width=\linewidth]{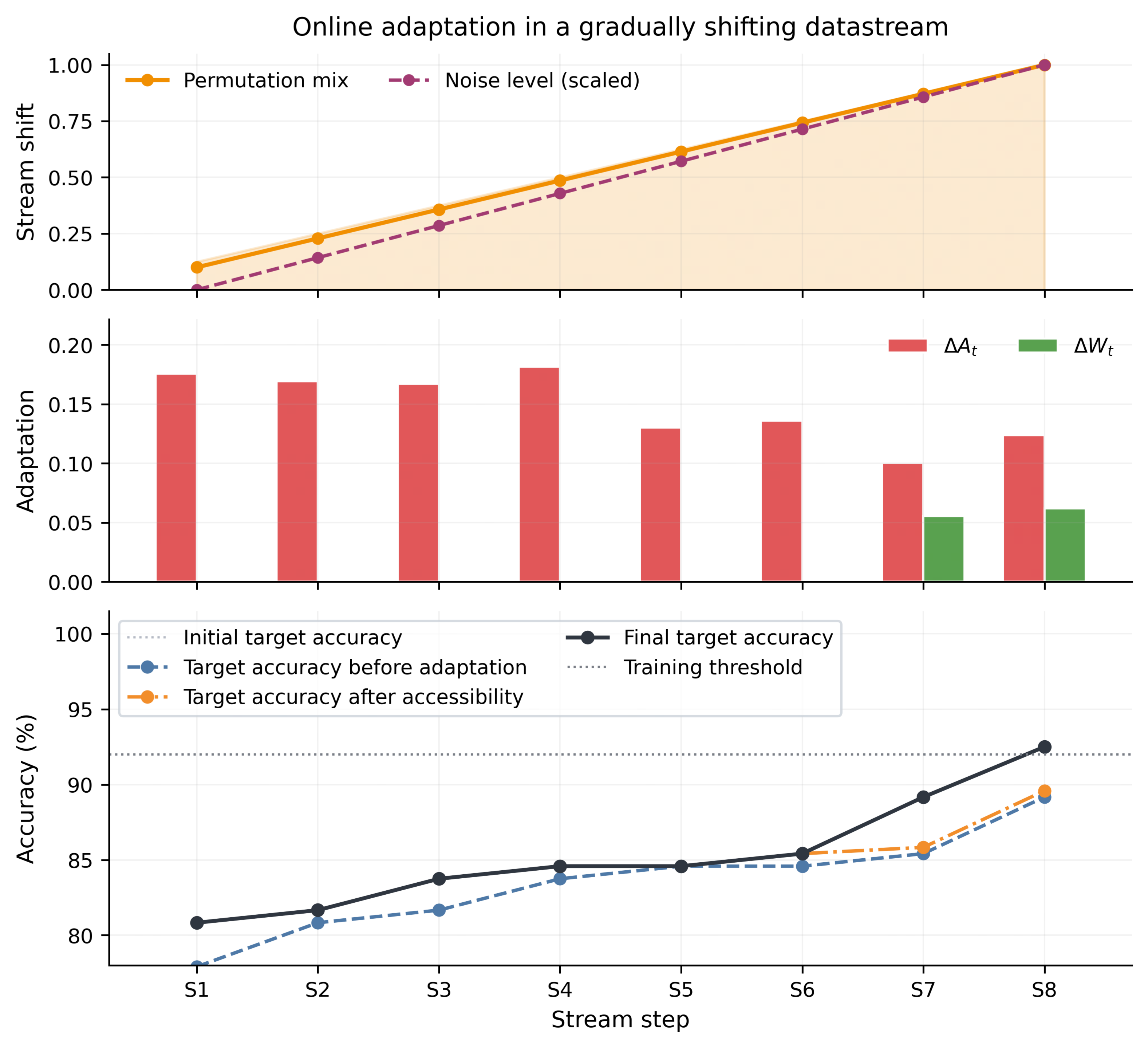}
\caption{
Online adaptation in a gradually shifting datastream. A modest initial state
is established first, after which the same binary MNIST classifier is exposed to
an eight-step stream with increasing permutation mix and noise level. The top
panel shows the controlled shift schedule, the middle panel reports accessibility
and capability modification at each step, and the bottom panel tracks accuracy on
the fixed hardest target environment before adaptation, after accessibility
adaptation, and after the final model state. The trajectory shows continuous
accessibility evolution throughout the stream, with capability plasticity
appearing only at the latest and hardest steps.
}
\label{fig:dynamic_accessibility}
\end{figure}

\subsubsection{Dynamic Accessibility Evolution}
Figure~\ref{fig:dynamic_accessibility} tests the temporal interpretation of
Accessibility Plasticity. Before the stream begins, the initialized model state
reaches $0.7792$ accuracy on the hardest target environment. As the datastream
progressively approaches that environment, target accuracy rises from $0.8083$ at
the first stream step to $0.9250$ at the last. Accessibility changes at every
step, with $\Delta A_t$ remaining nonzero throughout the stream, while shared
capability modification is withheld until the final two and hardest steps
($\Delta W_t=0.0556$ and $0.0620$). The stagewise curves show that accessibility
adaptation already improves target accuracy during the stream, from $0.7792$ to
$0.8958$ before the late capability updates, after which the final target
accuracy reaches $0.9250$. This experiment is intentionally small, but it
provides a clearer proof-of-concept for the claim that accessibility is not only
an optimization variable at an endpoint; it can function as a continuously
evolving adaptive state under feedback from a changing datastream.

\subsection{Interpretation Relative to the Central Claims}

The experiments support this claim. Removing accessibility adaptation while
keeping the shared capability bank produces a materially different regime.
Compared with Fixed Relationships, the operational realization reduces capability
modification by
approximately $58.0\%$ on Split MNIST and $64.0\%$ on Permuted MNIST. This is the
most direct evidence in the current experiments that accessibility is not a
decorative auxiliary variable, but a meaningful adaptive degree of freedom.

The support for this claim is positive but appropriately qualified. After the
first task, the operational realization follows the trace reuse $\rightarrow$ accessibility
$\rightarrow$ capability on every subsequent task in both datasets. Accessibility
adaptation often improves the current-task training signal before capability
plasticity is enabled, and the relationship state is frozen once capability
plasticity begins. Across the eight post-initial tasks, accessibility
adaptation improves current-task training accuracy in seven cases. The effect is
especially visible on Permuted MNIST, where the accessibility stage raises
current-task training accuracy from $0.520$ to $0.544$, from $0.618$ to $0.650$,
from $0.494$ to $0.576$, and from $0.564$ to $0.792$ before shared capability is
updated. However, accessibility adaptation alone is usually not sufficient to
reach the target threshold, so capability plasticity is still invoked. The
experiments therefore support a reduction-before-rewriting interpretation rather
than a claim that accessibility generally eliminates the need for capability
change.

The random accessibility control is designed precisely for this question. It
matches the operational realization at the level of accessibility-state budget, but replaces
relationship-driven updates with random perturbations. If the effect were only a
consequence of adding extra adaptable state, the random control should behave
similarly to the learned realization. It does not. On Split MNIST, the random control attains lower
accuracy ($0.8858$ versus $0.9183$) while spending much more accessibility cost
($2.7926$ versus $1.4131$). On Permuted MNIST, the gap is even clearer:
$0.8608$ versus $0.9242$ in accuracy, with $2.9599$ versus $0.5597$ in
accessibility cost. These comparisons support the claim that structured
relationship adaptation matters, not merely the presence of extra adaptive
parameters.

The Accessibility Only variant strongly supports this claim. Once later
capability plasticity is disabled, final accuracy drops to $0.6625$ on Split
MNIST and $0.6375$ on Permuted MNIST. Forgetting is low in this condition, but
that low forgetting is inseparable from the model's limited ability to absorb the
full task sequence. In other words, accessibility plasticity can carry part of the
adaptation burden, but not the whole burden. This is consistent with the proposed
hierarchy: reuse and accessibility plasticity are meaningful shallower responses,
but deeper capability plasticity remains necessary when the required computation
does not become sufficiently available through accessibility alone.

\subsection{Relationship to Existing Adaptive Mechanisms}

These experiments do not establish state-of-the-art continual learning, nor do
they show superiority over all continual-learning baselines. They also do not
constitute empirical comparisons against attention, routing, gating,
mixture-of-experts systems, dynamic networks, or deeper organizational plasticity. That is
intentional. The present paper proposes Accessibility Plasticity as a higher-level
adaptive abstraction rather than as a competing architectural mechanism. Existing
mechanisms such as attention, routing, gating, and mixture-of-experts may be
viewed as possible realizations of accessibility modulation at different levels of
description; direct empirical comparisons with those mechanisms are left for future
work.

\subsection{Limitations and Future Work}

The current empirical evidence should therefore be interpreted narrowly and
carefully. It provides proof-of-concept support for the claim that accessibility
is a distinct computational object whose adaptation can precede and reduce shared
capability modification. It does not yet validate the deepest level of the
hierarchy, $\Delta(D,G)$, and it does not yet provide the scale of evidence that
would be required for a mature benchmark paper. Natural next steps include
multiple random seeds, confidence intervals, larger continual-learning benchmarks,
stronger ablation studies, larger-scale online environments, embodied agents, and
lifelong-learning settings in which environmental change is not organized into
hand-designed phases. These extensions would broaden the empirical case; they are
not prerequisites for the narrower proof-of-concept conclusions drawn here.

\section{Conclusion}

This paper develops Accessibility Plasticity as a paradigm for reuse-first
adaptation and presents one graph-based operational realization. The key move is to separate computational capability from
computational accessibility and to treat accessibility as a fundamental
computational object. In this view, $W$ determines what computation exists,
whereas $\mathcal{A}$ determines what computation is accessible. Learning is
therefore not only changing what computation exists; it is changing how existing
computation becomes accessible. The resulting framework is mathematically simple
enough to instantiate, yet conceptually broad enough to connect continual learning
and adaptive computation under a shared principle.

If validated empirically, the paradigm would support a shift from
architecture-centric learning toward relationship-centric adaptation. Even if future
experiments reveal limitations, the formulation provides a rigorous starting point
for studying whether intelligent systems can adapt by reorganizing computation
before rebuilding it.

\bibliographystyle{plain}
\bibliography{references}

\end{document}